\documentclass{article}

\usepackage[dblblindworkshop,final]{neurips_2026}
\workshoptitle{Global South AI: Rethinking AI for and from the Global South}

\usepackage[utf8]{inputenc}
\usepackage[T1]{fontenc}
\usepackage{hyperref}
\usepackage{url}
\usepackage{booktabs}
\usepackage{amsfonts}
\usepackage{amsmath}
\usepackage{nicefrac}
\usepackage{microtype}
\usepackage{xcolor}
\usepackage{graphicx}
\usepackage{placeins}

\hypersetup{pdfauthor={Chibueze Praise Onyenweaku, Toyibat Adele, Sikirulahi Opeyemi Abdulkareem, Obed Honour Eje},
            pdftitle={Occlusion for Free: Detecting Occlusion from the Disagreement Between Two Monocular Distance Cues}}

\title{Occlusion for Free:\\
       Detecting Occlusion from the Disagreement\\
       Between Two Monocular Distance Cues}

\author{%
  Chibueze Praise Onyenweaku$^{1}$ \quad Toyibat Adele$^{2,*}$ \\
  \textbf{Sikirulahi Opeyemi Abdulkareem$^{2,*}$ \quad Obed Honour Eje$^{3,*}$} \\[0.5ex]
  \normalfont $^{1}$University of Lagos \quad
  $^{2}$ML Collective \quad
  $^{3}$Federal University of Technology, Minna \\
}

\begin{document}

\maketitle

\makeatletter
\def\blfootnote{\gdef\@thefnmark{}\@footnotetext}
\makeatother

\blfootnote{$^*$ Equal contribution.}

\begin{abstract}
Knowing \emph{that} a pedestrian is occluded is useful to a robot even when the detector cannot
recover the hidden pixels, but occlusion estimators are normally trained on occlusion annotations
that deployment settings rarely have. We show that occlusion can be read off two distance estimates
a camera-only pipeline already computes. Apparent height is largely unchanged by occlusion, because
the detected box still spans the whole person; monocular metric depth is pulled toward the occluder,
because the nearest surface inside the box is what a depth map reports. Only their ordering matters,
so no focal length, stature, or depth scale is needed, provided the depth model's error is close to a
pure gain, a condition that can be checked without labels. The score detects heavy occlusion on KITTI
with AUC 0.909 (0.888 over all annotated pedestrians, including those the detector misses), requires
no training and no occlusion labels, and is not a proxy for range: within fixed distance bands it
holds AUC 0.894--0.925 while true distance alone sits near chance. On synthetic CARLA scenes with
measured visibility it reaches 0.819, against 0.644 for distance alone. We arrived here after
measuring three standard architectural responses to occlusion, depth fusion, occlusion-simulating
augmentation, and attention, none of which improved detection outside its training distribution in
the configurations we tested, and after finding that a detector trained on German roads loses half
its accuracy on self-annotated street scenes in Lagos, Nigeria. Signals that need no local
annotation are worth more in settings that benchmarks do not represent.
\end{abstract}


\section{Introduction}

Robots need to know what is around them, and a camera is the cheapest way to sense it. People and
objects are often only partly visible, and this occlusion is the dominant failure condition for
pedestrian detection in cluttered environments [1]. A robot does not
always need to recover the hidden pixels. Often it is enough to know \emph{that} a detection is
occluded, so it can widen an uncertainty estimate or shorten a planning horizon.

Most prior work addresses occlusion by changing the detector, fusing separately estimated depth with
the RGB image [2--4], simulating occlusion during training by masking parts of objects [5--8], or
adding attention that re-weights features toward visible parts [9,10]. Designing
for a low-cost robot, we measured all three; none improved detection outside its training
distribution in the configurations we tested, and depth fusion hurt accuracy even in-distribution,
matching a known limit of monocular depth models [11,12]. A detector trained on this benchmark data
also lost about half its accuracy on 350 self-annotated Lagos frames.

Our depth branch failed as a detection input, but that failure is what our approach is built on. A
depth map reports how far away the nearest visible surface is at each point. Inside an occluded
pedestrian's box, that nearest surface is usually the occluder, not the person
behind it, so the depth value read is pulled toward the occluder's distance. Apparent
height is unaffected, since the box still spans the whole person. The two cues
move in \emph{opposite} directions under occlusion, so their disagreement is a direct readout of it,
computed from two numbers the pipeline already produces. \textbf{Contributions.} A training-free,
label-free occlusion signal from this disagreement, validated with a distance-band check that rules
out range as a confound and with tests of each link in its mechanism; and a controlled comparison of
depth fusion, augmentation, and attention across 81 checkpoints, together with a self-collected Lagos
evaluation set measuring the detector transfer gap that motivates a training-free signal.

\section{Related Work}

\textbf{Depth without LiDAR.} MV3D-style LiDAR fusion [2] remains the accuracy reference but is hard
to justify on low-cost platforms. MiDaS [3] predicts only relative depth, with no fixed scale;
Depth Anything V2 [4] recovers absolute scale, and we use its outdoor variant, fine-tuned on
synthetic Virtual KITTI rather than LiDAR-supervised frames, so evaluating on KITTI leaks no measured
depth. Monocular depth decoders also blur values across occlusion boundaries, since a convolution
kernel mixes information across surfaces in its receptive field [11], and a concurrent study
reports depth can make an RGB-D detector perform worse than RGB alone [12].

\textbf{Occlusion robustness vs.\ occlusion estimation.} Augmentation,
single-region masking [5,6], multi-patch hiding [7], grid erasure [8], and attention that
reweights features toward occluded pedestrians [9,10] all try to make a detector perform
better despite occlusion. Neither tells you whether a specific detection is actually occluded.
Stereo methods detect occlusion directly from left--right disparity inconsistency [13]. MonoLoco
[14] comes closest, but treats wider uncertainty on occluded cases as incidental rather than
validated, and never checks whether the effect is occlusion or just distance. Bi-box regression also
estimates occlusion but needs paired visible and full-body labels [15]. We instead treat the
disagreement between two distance estimates as the signal itself, training-free and label-free.

\section{Method}

\subsection{System Overview}

\begin{figure}[htbp]
  \centering
  \includegraphics[width=\linewidth]{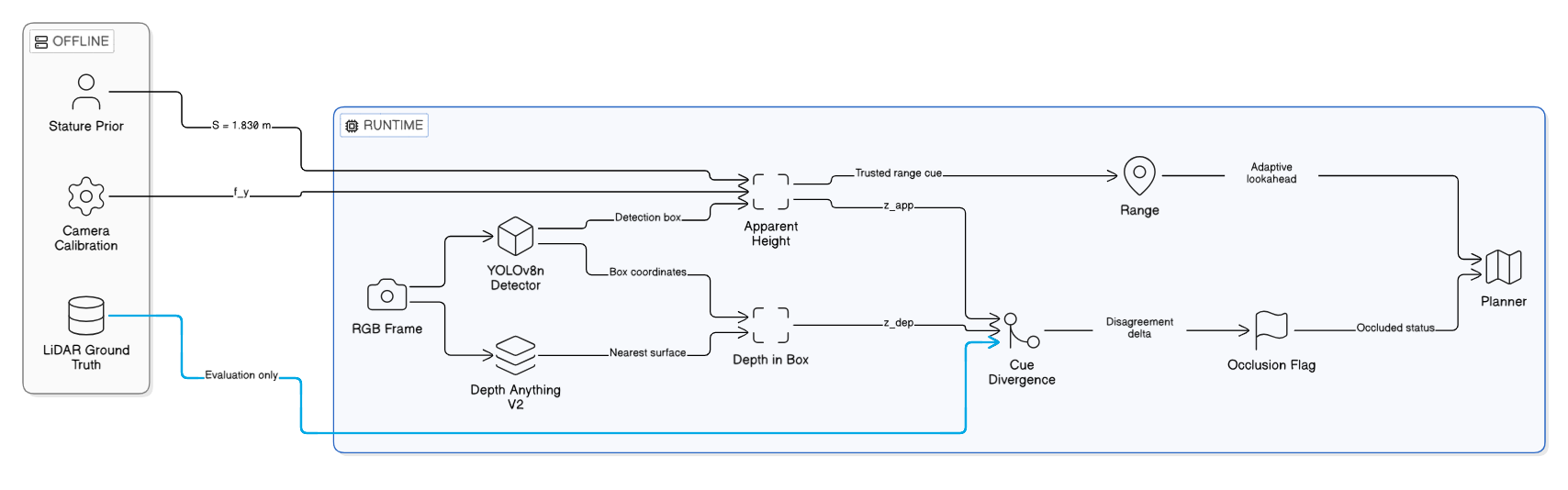}
  \caption{Runtime pipeline. Offline inputs are fixed before deployment.
  At runtime, two monocular distance estimates per detection are compared;
  their disagreement flags occlusion without training or occlusion labels.
  LiDAR ground truth is used for evaluation only, and stature and focal
  length enter only metric read-outs, not the ranking score.}
  \label{fig:system}
\end{figure}

Figure~\ref{fig:system} shows the full pipeline. An RGB frame enters a
YOLOv8 detector [16,17] (YOLOv8x in our evaluation), producing bounding boxes
for each detected pedestrian. Those boxes feed two parallel distance
estimators, an apparent-height module and a Depth Anything V2~[4] branch
reading the nearest surface inside each box. Their disagreement produces a
scalar occlusion score passed to the planner. LiDAR ground truth is used only
for evaluation; the runtime path is entirely camera-based. We arrived at this
design after the three model-side responses to occlusion, measured across 81
checkpoints (Section~\ref{sec:elim}), failed to transfer in the configurations
tested, which turned the
question toward \emph{why} depth failed as a detector input.

\subsection{Monocular Distance Cues}

\subsubsection{Apparent Height Estimation}

The apparent-height distance estimate is obtained by inverting the pinhole camera relationship, $z_\mathrm{app} = f_y S / h_\mathrm{px}$, where $f_y$ is the vertical focal length of the camera, $S$ is the effective stature of the pedestrian, and $h_\mathrm{px}$ is the height of the pedestrian's bounding box in pixels. $S = 1.83$\,m is the median implied stature of fully-visible annotated KITTI pedestrians; it cancels in the ranking score below.

KITTI boxes span the pedestrian's full extent even when part of it is occluded, so $z_\mathrm{app}$ is insensitive to occlusion as long as the box does not shrink toward the visible part. The score uses \emph{predicted} boxes, so we test this on detector output: the implied stature from YOLOv8x boxes is 1.825, 1.811 and 1.813\,m for fully-visible, partly and largely occluded pedestrians, though a tail of low-confidence occluded boxes is short (Appendix~\ref{app:mechanism}).

\subsubsection{Monocular Metric Depth Estimation}

The second distance cue is obtained using a frozen Depth Anything V2 outdoor metric depth model. For each detected pedestrian, $z_\mathrm{dep} = k \cdot d_\mathrm{min}$, where $d_\mathrm{min}$ is the minimum depth value in the box and $k$ is a scale factor fitted on fully-visible instances, used only by the calibrated read-outs of Appendix~\ref{app:formulation}. Unlike apparent height, $z_\mathrm{dep}$ is sensitive to occlusion, since a closer occluder's surface can dominate that minimum, pulling the estimate toward the camera. Low percentiles, robust to a single erroneous pixel, perform as well or better (Appendix~\ref{app:mechanism}); our released implementation uses the 5th percentile.

\subsection{Occlusion Score Formulation}

Because the two cues respond differently to occlusion, apparent height stable, depth pulled toward the occluder, their relationship is itself an indicator of occlusion. The ratio between the depth-based and apparent-height distance estimates can be written as

\begin{equation}
  \frac{z_\mathrm{dep}}{z_\mathrm{app}}
    \;=\; \underbrace{\frac{k}{f_y S}}_{\textstyle G} \; d_\mathrm{min} \, h_\mathrm{px},
  \label{eq:score}
\end{equation}

where all calibration-related terms are combined into a single positive gain $G$.

We only need to \emph{rank} detections by likelihood of occlusion, and multiplying by a positive constant does not change a ranking, so $G$ can be dropped. The deployable score is $d_\mathrm{min} h_\mathrm{px}$, requiring no focal length, stature, or depth rescaling factor; lower values indicate occlusion, since a closer occluder reduces the minimum depth inside the box.

This invariance is to a depth \emph{gain}, not an \emph{offset}, and it requires metric depth. If the depth model reads $d_\mathrm{min} = a\,z + b$, then for an unoccluded pedestrian, where $z_\mathrm{app} \approx z$,
\begin{equation}
  d_\mathrm{min}\, h_\mathrm{px} \;\approx\; f_y S \left(a + \frac{b}{z_\mathrm{app}}\right).
  \label{eq:affine}
\end{equation}
The gain $a$ cancels in the ranking; the offset $b$ adds a term that falls with distance, turning the score partly into a range measure. This is checkable without occlusion labels, because an offset makes the score correlate with distance among visible pedestrians: for Depth Anything V2's outdoor checkpoint on KITTI, $\rho = -0.07$, and a synthetic $+2$\,m offset moves it to $-0.61$ (Appendix~\ref{app:formulation}).

\section{Experiments}

\subsection{Setup}

We train YOLOv8n and YOLOv8x on KITTI [18] and CityPersons [19] using byte-identical settings and standard train/validation/test splits. Three seeds are used where confidence intervals are reported and one where marked, giving 81 evaluated checkpoints. Occlusion augmentation is applied only during training and only when more than half of the ground-truth pedestrian remains visible. We report mAP@0.5 by KITTI occlusion level.

KITTI labels pedestrians fully visible (0), partly (1) or largely occluded (2), or unknown (3); \emph{heavy} is 2--3 and \emph{any} is 1--3, each against visible. The score is evaluated on 1{,}941 detections from three YOLOv8x seeds, matched to 682 distinct test pedestrians by greedy IoU~$\geq$~0.5. AUC measures whether an occluded pedestrian is ranked above a fully-visible one; since a pedestrian contributes up to three rows, 95\% intervals bootstrap pedestrians (2{,}000 resamples).

\subsection{Results}

\begin{figure}[htbp]
  \centering
  \includegraphics[width=\linewidth]{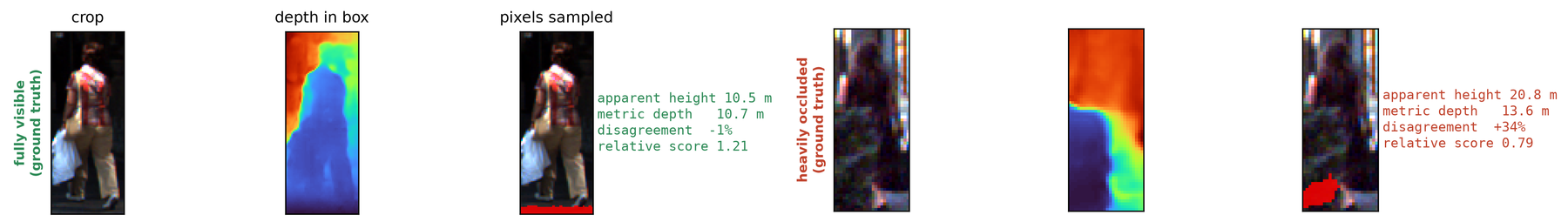}
  \caption{One KITTI frame showing the proposed cue. The visible pedestrian's nearest surface is their own and the two distance estimates agree; for the occluded pedestrian, depth reads the nearer occluder and the estimates diverge by 34\%.}
  \label{fig:qualitative}
\end{figure}

Figure~\ref{fig:qualitative} shows the mechanism on one KITTI frame. For the visible pedestrian, apparent height and depth agree (10.5\,m vs.\ 10.7\,m). For the occluded pedestrian, the nearest surface belongs to the occluder, apparent height gives 20.8\,m while depth gives 13.6\,m. The resulting 34\% disagreement reduces the score to 0.79 of the scene median.

\subsection{Occlusion Detection and Distance Control}

The score reaches AUC 0.909 for heavy occlusion and 0.826 for any occlusion on matched detections, and 0.775 for partly occluded pedestrians alone. The detector misses 27\% of largely occluded pedestrians (3\% of visible ones), so matched detections favor the easier occluded cases; scoring all 709 annotated test pedestrians with annotated boxes, no detector involved, gives 0.888 [0.852, 0.921].

A distance confound is unlikely. Occluded pedestrians are farther away on average (16.9\,m vs.\ 13.2\,m), but within fixed distance bands true distance falls to near chance (AUC 0.457--0.594), while the proposed score remains at 0.894--0.925 for heavy occlusion (Table~\ref{tab:auc}). The 5--10\,m band rests on only 10 distinct heavily occluded pedestrians, so its interval is wide. The full breakdown, for any, partial and heavy occlusion, is in Appendix~\ref{app:distance}.

\begin{table}[htbp]
\caption{Occlusion score on KITTI test, heavy occlusion vs.\ fully-visible. $N$ counts matched detections, with distinct pedestrians in parentheses; intervals bootstrap pedestrians. The final column is the AUC of true distance alone on the same instances.}
\label{tab:auc}
\centering
\small
\begin{tabular}{lccc}
\toprule
Distance band & $N$ vis/occ (peds) & AUC of score [95\% CI] & AUC of distance \\
\midrule
5--10\,m  & 409/26 (137/10)    & 0.917 [0.843, 0.971] & 0.594 \\
10--20\,m & 466/140 (156/55)   & 0.894 [0.840, 0.939] & 0.457 \\
20\,m$+$  & 318/100 (112/41)   & 0.925 [0.874, 0.964] & 0.469 \\
all       & 1237/269 (420/107) & \textbf{0.909 [0.876, 0.938]} & 0.624 \\
\bottomrule
\end{tabular}
\end{table}

\subsection{Formulation, Depth Source and Mechanism}

The calibration-free ranking performs identically to calibrated alternatives (0.909 vs.\ 0.908) and requires metric rather than ordinal depth: MiDaS's relative output reverses the ranking (Appendix~\ref{app:formulation}). The result survives each test of its mechanism we could construct (Appendix~\ref{app:mechanism}): swapping predicted and annotated boxes for either cue moves heavy AUC by at most 0.010, low percentiles in place of the minimum give 0.910--0.921, and box shifts of up to 4\,px give 0.904--0.907.

\subsection{Beyond KITTI, and Runtime}

On 573 synthetic CARLA frames (3{,}293 pedestrians), with visibility measured by re-rendering without occluders, the score reaches AUC 0.819 [0.801, 0.834] against 0.644 for distance alone, beating distance in every band (Appendix~\ref{app:carla}). CARLA's boxes are oracle and its depth checkpoint was chosen on CARLA ground-truth depth, so this tests the depth mechanism under a change of domain and camera height, not the full pipeline. The score itself costs about 0.3\,ms per frame on a laptop CPU; the depth network dominates (Appendix~\ref{app:runtime}).

\subsection{Model-Side Alternatives and Lagos Transfer}
\label{sec:elim}

Table~\ref{tab:elim} (Appendix~\ref{app:elim}) reports three model-side interventions. With three YOLOv8x seeds, augmentation improves hard KITTI cases by $+0.041$ mAP50 but not CityPersons, and MiDaS depth fusion reduces accuracy. The DAv2 fusion and POPAM rows are single YOLOv8n runs; they record what those configurations did, not that the interventions generally fail.

On 350 self-annotated Lagos frames (Table~\ref{tab:transfer}), a KITTI-trained detector drops from 0.8809 to 0.4482 mAP50, while the CityPersons-trained baseline reaches 0.7356, a $+0.2873$ difference on identical frames. The datasets differ in size, density, diversity, geography, camera and annotation protocol, so this shows only that the CityPersons-trained model transfers better. It evaluates the detector, not the score: the Lagos annotations have no occlusion labels.

\begin{ack}
We thank OpenToken and CloudRift for providing the GPU compute used for this work.

\end{ack}

\FloatBarrier

\section*{References}

{
\small

[1] Z. Ouardirhi, S. A. Mahmoudi, and M. Zbakh, ``Enhancing Object Detection in Smart Video Surveillance: A Survey of Occlusion-Handling Approaches,'' \textit{Electronics}, vol. 13, no. 3, p. 541, 2024. \url{https://doi.org/10.3390/electronics13030541}

[2] X. Chen, H. Ma, J. Wan, B. Li, and T. Xia, ``Multi-View 3D Object Detection Network for Autonomous Driving,'' in \textit{Proc. IEEE CVPR}, 2017. \url{https://arxiv.org/abs/1611.07759}

[3] R. Ranftl, K. Lasinger, D. Hafner, K. Schindler, and V. Koltun, ``Towards Robust Monocular Depth Estimation: Mixing Datasets for Zero-shot Cross-dataset Transfer,'' \textit{IEEE TPAMI}, vol. 44, no. 3, pp. 1623--1637, 2022. \url{https://doi.org/10.1109/TPAMI.2020.3019967}

[4] L. Yang, B. Kang, Z. Huang, Z. Zhao, X. Xu, J. Feng, and H. Zhao, ``Depth Anything V2,'' in \textit{NeurIPS}, 2024. \url{https://arxiv.org/abs/2406.09414}

[5] T. DeVries and G. W. Taylor, ``Improved Regularization of Convolutional Neural Networks with Cutout,'' 2017. \url{https://arxiv.org/abs/1708.04552}

[6] Z. Zhong, L. Zheng, G. Kang, S. Li, and Y. Yang, ``Random Erasing Data Augmentation,'' in \textit{AAAI}, vol. 34, no. 7, 2020, pp. 13001--13008. \url{https://arxiv.org/abs/1708.04896}

[7] K. K. Singh, H. Yu, A. Sarmasi, G. Pradeep, and Y. J. Lee, ``Hide-and-Seek: A Data Augmentation Technique for Weakly-Supervised Localization and Beyond,'' 2018. \url{https://arxiv.org/abs/1811.02545}

[8] P. Chen, S. Liu, H. Zhao, X. Wang, and J. Jia, ``GridMask Data Augmentation,'' 2020. \url{https://arxiv.org/abs/2001.04086}

[9] H. Liu, Z. Zhang, and B. Feng, ``POP-YOLOv8: An Object Detection Framework for Partially Occluded Pedestrians in Nighttime Traffic Environments,'' \textit{Scientific Reports}, vol. 16, no. 1, p. 4841, 2026. \url{https://www.nature.com/articles/s41598-026-35146-9}

[10] S. Tang, Y. Zhou, J. Li, C. Liu, and J. Shi, ``Attention-Guided Sample-Based Feature Enhancement Network for Crowded Pedestrian Detection Using Vision Sensors,'' \textit{Sensors}, vol. 24, no. 19, art. 6350, 2024. \url{https://doi.org/10.3390/s24196350}

[11] M. Ramamonjisoa, Y. Du, and V. Lepetit, ``Predicting Sharp and Accurate Occlusion Boundaries in Monocular Depth Estimation Using Displacement Fields,'' in \textit{Proc. IEEE/CVF CVPR}, 2020. \url{https://arxiv.org/abs/2002.12730}

[12] X. Wang, J. Hua, R. Li, Z. Wu, C. Chen, K. Gu, and A. Hao, ``When Depth Hurts: Reliability-Aware Geometry Distillation for Depth-Free RGB-D Salient Object Detection,'' 2026. \url{https://arxiv.org/abs/2609.03378}

[13] G. Egnal and R. P. Wildes, ``Detecting Binocular Half-Occlusions: Empirical Comparisons of Five Approaches,'' \textit{IEEE TPAMI}, vol. 24, no. 8, pp. 1127--1133, 2002. doi: 10.1109/TPAMI.2002.1023808

[14] L. Bertoni, S. Kreiss, and A. Alahi, ``MonoLoco: Monocular 3D Pedestrian Localization and Uncertainty Estimation,'' in \textit{Proc. IEEE/CVF International Conference on Computer Vision (ICCV)}, 2019. \url{https://arxiv.org/abs/1906.06059}

[15] C. Zhou and J. Yuan, ``Bi-box Regression for Pedestrian Detection and Occlusion Estimation,'' in \textit{Proc. European Conference on Computer Vision (ECCV)}, 2018, pp. 135--151. doi: 10.1007/978-3-030-01246-5\_9

[16] J. Redmon, S. Divvala, R. Girshick, and A. Farhadi, ``You Only Look Once: Unified, Real-Time Object Detection,'' in \textit{Proc. IEEE CVPR}, 2016, pp. 779--788. doi: 10.1109/CVPR.2016.91

[17] Ultralytics, ``YOLOv8 Documentation, Performance Metrics,'' 2024. \url{https://docs.ultralytics.com/models/yolov8/}

[18] A. Geiger, P. Lenz, and R. Urtasun, ``Are we ready for Autonomous Driving? The KITTI Vision Benchmark Suite,'' in \textit{Proc. IEEE CVPR}, 2012, pp. 3354--3361. doi: 10.1109/CVPR.2012.6248074

[19] S. Zhang, R. Benenson, and B. Schiele, ``CityPersons: A Diverse Dataset for Pedestrian Detection,'' in \textit{Proc. IEEE CVPR}, 2017, pp. 3213--3221. \url{https://arxiv.org/abs/1702.05693}

}

\appendix

\section{Full Distance-Band and Per-Level Breakdown}
\label{app:distance}

Table~\ref{tab:auc-full} gives the full distance-band breakdown that Table~\ref{tab:auc} summarizes, for any occlusion, for partly occluded pedestrians (level 1) alone, and for heavy occlusion. The score beats true distance in every band for every target. It is weakest for partial occlusion at close and middle range (0.728--0.735) and strongest for heavy occlusion. Computed separately for each detector seed, heavy and any AUC are $0.909\pm0.007$ and $0.826\pm0.003$.

\begin{table}[htbp]
\caption{Occlusion score on KITTI test versus fully-visible pedestrians. $N$ counts matched detections, with distinct pedestrians in parentheses; intervals bootstrap pedestrians. The final column is the AUC of true distance alone on the same instances.}
\label{tab:auc-full}
\centering
\small
\begin{tabular}{llccc}
\toprule
Target & Band & $N$ vis/occ (peds) & AUC of score [95\% CI] & AUC of distance \\
\midrule
Any (1--3)   & 5--10\,m  & 409/124 (137/44)   & 0.773 [0.676, 0.863] & 0.549 \\
Any (1--3)   & 10--20\,m & 466/297 (156/109)  & 0.807 [0.750, 0.860] & 0.471 \\
Any (1--3)   & 20\,m$+$  & 318/267 (112/102)  & 0.874 [0.828, 0.919] & 0.513 \\
Any (1--3)   & all       & 1237/704 (420/262) & 0.826 [0.792, 0.857] & 0.600 \\
\midrule
Partly (1)   & 5--10\,m  & 409/98 (137/34)    & 0.735 [0.615, 0.840] & 0.538 \\
Partly (1)   & 10--20\,m & 466/157 (156/54)   & 0.728 [0.648, 0.809] & 0.485 \\
Partly (1)   & 20\,m$+$  & 318/167 (112/61)   & 0.844 [0.781, 0.900] & 0.539 \\
Partly (1)   & all       & 1237/435 (420/155) & 0.775 [0.728, 0.817] & 0.585 \\
\midrule
Heavy (2--3) & 5--10\,m  & 409/26 (137/10)    & 0.917 [0.843, 0.971] & 0.594 \\
Heavy (2--3) & 10--20\,m & 466/140 (156/55)   & 0.894 [0.840, 0.939] & 0.457 \\
Heavy (2--3) & 20\,m$+$  & 318/100 (112/41)   & 0.925 [0.874, 0.964] & 0.469 \\
Heavy (2--3) & all       & 1237/269 (420/107) & \textbf{0.909 [0.876, 0.938]} & 0.624 \\
\bottomrule
\end{tabular}
\end{table}

\section{Score Formulation, Depth Source, and Depth Offset}
\label{app:formulation}

Table~\ref{tab:form-full} compares three read-outs of the cue disagreement, referenced in Section~4.4. The ranking score and the signed ratio are monotone and perform identically, so removing $f_y$, $S$, and $k$ costs nothing. The absolute difference performs worse and collapses once the depth scale is perturbed 25\% high, confirming that ranking, not magnitude, is what survives calibration drift. Depth reads farther than apparent height for 44\% of fully-visible pedestrians but only 3\% of heavily occluded ones, so the direction of the disagreement is itself informative and taking its magnitude discards that.

\begin{table}[htbp]
\caption{Occlusion AUC for three formulations of the cue disagreement. Heavy occlusion.}
\label{tab:form-full}
\centering
\small
\begin{tabular}{lcc}
\toprule
Read-out & Calib. & AUC \\
\midrule
Ranking, $d_\mathrm{min} h_\mathrm{px}$ & none & \textbf{0.909} \\
Signed, $1-z_\mathrm{dep}/z_\mathrm{app}$ & $f_y,S,k$ & 0.908 \\
Absolute difference & $f_y,S,k$ & 0.872 \\
Absolute, gain 25\% high & $f_y,S,k$ & 0.470 \\
\bottomrule
\end{tabular}
\end{table}

Table~\ref{tab:depth-full} compares depth sources. MiDaS used as-is reverses the ordering, while inverting it recovers some performance but remains below Depth Anything V2. The signal therefore needs depth proportional to distance, not simply an arbitrary monocular depth ordering.

\begin{table}[htbp]
\caption{Occlusion AUC by depth source.}
\label{tab:depth-full}
\centering
\small
\begin{tabular}{lcc}
\toprule
Depth source & any & heavy \\
\midrule
MiDaS, as-is    & 0.572 & 0.596 \\
MiDaS, inverted & 0.674 & 0.798 \\
DAv2, metric    & \textbf{0.826} & \textbf{0.909} \\
\bottomrule
\end{tabular}
\end{table}

\textbf{Depth offset.} Eq.~\ref{eq:affine} predicts that an additive depth offset $b$ turns the score partly into a range measure. Adding a synthetic offset to $d_\mathrm{min}$ reproduces this. At $b=0$ the score is nearly uncorrelated with distance among visible pedestrians ($\rho=-0.07$). At $b=+2$ and $+4$\,m, $\rho=-0.61$ and $-0.81$, and the weakest within-band heavy AUC falls from 0.894 to 0.839 and 0.800. Overall AUC falls more slowly, to 0.876 at $+4$\,m, which is why within-band AUC, not overall AUC, is the criterion to report. The correlation needs no occlusion labels, so it is a deployment check for any new depth checkpoint or camera mounting. The calibration-free property holds on KITTI because this checkpoint's offset there is negligible. It is not a general guarantee: on CARLA (Appendix~\ref{app:carla}) the depth checkpoint used has a substantial offset.

\section{Mechanism Checks}
\label{app:mechanism}

A high AUC does not by itself show that the mechanism in the Introduction produces it. Table~\ref{tab:mechanism} tests each link on the same KITTI detections and depth maps.

\textbf{Detector boxes.} If occluded boxes shrank, $z_\mathrm{app}$ would inflate and the score could read localization error. The median predicted-to-annotated box-height ratio is 0.99--1.00 at every occlusion level, but 15\% of largely occluded predicted boxes are more than 10\% short (1\% of visible), mostly low-confidence detections. Swapping either ingredient between predicted and annotated boxes changes heavy AUC by at most 0.010. Scoring all 709 annotated pedestrians with annotated boxes, no detector involved, gives 0.888. It is lower than 0.909 because the detector misses 27\% of largely occluded pedestrians (10\% of partly occluded, 3\% of visible), so matched detections favor the easier occluded cases.

\textbf{The minimum operator.} Low percentiles perform as well as the minimum or better (0.910--0.921), and the signal fades only toward the median (0.859), where the occluder no longer dominates. Shrinking the box helps, growing it admits ground and background, and shifts of up to 4\,px barely matter: the near part of the depth distribution carries the effect, not one pixel.

\textbf{Other confounds.} Excluding truncated pedestrians raises heavy AUC to 0.913. Stature varies with log-s.d.\ 0.064 among visible pedestrians, against a median log-score shift of 0.37 under heavy occlusion. The score is weaker on low-confidence detections (0.867 in the lowest confidence tercile, 0.903 in the middle; the top tercile has too few heavy cases to score). Pose is unlabeled in KITTI and untested.

\begin{table}[htbp]
\caption{Mechanism checks, KITTI heavy occlusion vs.\ visible. Pedestrian-bootstrap 95\% intervals.}
\label{tab:mechanism}
\centering
\small
\begin{tabular}{lc}
\toprule
Variant & Heavy AUC [95\% CI] \\
\midrule
Paper score, $d_\mathrm{min}$ in predicted box & 0.909 [0.876, 0.938] \\
$h$ annotated, $d_\mathrm{min}$ predicted box & 0.899 [0.866, 0.928] \\
$h$ predicted, $d_\mathrm{min}$ annotated box & 0.902 [0.868, 0.931] \\
Annotated boxes only, matched pedestrians & 0.904 [0.870, 0.931] \\
Annotated boxes, all 709 pedestrians & 0.888 [0.852, 0.921] \\
\midrule
1st / 5th / 10th / 25th percentile instead of min & 0.911 / 0.919 / 0.921 / 0.910 \\
Median & 0.859 [0.809, 0.902] \\
Box shrunk 10\% / grown 10\% & 0.921 / 0.892 \\
Box shifted $\leq$4\,px (5 draws) & 0.904--0.907 \\
\midrule
Truncation $\leq$ 0.1 only & 0.913 [0.882, 0.942] \\
Synthetic depth offset $b=+2$\,m / $+4$\,m & 0.900 / 0.876 \\
\bottomrule
\end{tabular}
\end{table}

\section{Model-Side Alternatives and Lagos Transfer, Full Results}
\label{app:elim}

Table~\ref{tab:elim} and Table~\ref{tab:transfer} give the full per-configuration results referenced in Section~\ref{sec:elim}. The 81 evaluated checkpoints are 39 KITTI YOLOv8x (three seeds: baseline, 7 augmentation and 5 MiDaS-fusion configurations, with and without augmentation), 15 KITTI YOLOv8n (one seed: baseline, 8 augmentation, 2 MiDaS-fusion, 1 DAv2-fusion and 3 POPAM configurations), and 27 CityPersons YOLOv8x (three seeds: baseline, 7 augmentation and 1 MiDaS-fusion configuration). Smoke tests, unevaluated architectures and earlier unseeded duplicates are excluded.

\begin{table}[htbp]
\caption{Three model-side interventions against their RGB-only baselines. Pedestrian class; $\Delta$ is mAP50 relative to baseline. $\dagger$ denotes single-seed results, which record what that configuration did rather than a general conclusion.}
\label{tab:elim}
\centering
\small
\begin{tabular}{llcc}
\toprule
Intervention & Setting & $\Delta$ mAP50 & $\Delta$ Hard mAP50 \\
\midrule
Occlusion augmentation (GridMask) & KITTI, YOLOv8x       & $+0.009$ & $+0.041$ \\
Occlusion augmentation (H\&S)     & CityPersons, YOLOv8x & $+0.001$ & $-0.004$ \\
Depth fusion (MiDaS)              & KITTI, YOLOv8x       & $-0.023$ & $-0.012$ \\
Depth fusion (MiDaS)$^\dagger$    & KITTI, YOLOv8n       & $-0.075$ & $-0.143$ \\
Depth fusion (DAv2)$^\dagger$     & KITTI, YOLOv8n       & $-0.056$ & $-0.093$ \\
Attention (POPAM)$^\dagger$       & KITTI, YOLOv8n       & $-0.030$ & $-0.055$ \\
\bottomrule
\end{tabular}
\end{table}

\begin{table}[htbp]
\caption{Detector transfer to 350 self-annotated Lagos frames (1{,}290 pedestrian boxes), mean $\pm$ s.d.\ over three training seeds. No model has seen this data. The annotations carry no occlusion labels, so the occlusion score is not evaluated here.}
\label{tab:transfer}
\centering
\small
\begin{tabular}{llcc}
\toprule
Trained on & Configuration & In-dist.\ mAP50 & Lagos mAP50 \\
\midrule
KITTI       & Baseline      & 0.8809 & 0.4482 $\pm$ 0.0094 \\
KITTI       & GridMask      & 0.8903 & 0.4150 $\pm$ 0.0530 \\
CityPersons & Baseline      & 0.7154 & \textbf{0.7356 $\pm$ 0.0351} \\
CityPersons & Hide-and-Seek & 0.7166 & 0.7075 $\pm$ 0.0198 \\
\bottomrule
\end{tabular}
\end{table}

\FloatBarrier

\section{Out-of-Domain Validation on CARLA}
\label{app:carla}

\textbf{Data and labels.} We render 573 frames in CARLA 0.9.16 (Towns 01--05) from a simulated low-mounted robot camera (0.152\,m above the ground, 1280$\times$720), giving 3{,}293 pedestrians. Every frame is rendered twice, once with occluding actors removed; a pedestrian's visibility is the ratio of their instance-mask pixel counts between the two renders, so labels are measured rather than annotated. A pedestrian with visibility below 0.8 is occluded (1{,}493 occluded, 1{,}800 visible). Depth is Depth Anything V2 Metric-Indoor-Large from RGB alone.

\textbf{What is and is not held out.} The score formulation, the use of full-extent boxes, and the 5th-percentile depth statistic were fixed from KITTI before CARLA was evaluated, and no occlusion label was used to choose anything. The depth checkpoint was chosen using CARLA data: the outdoor checkpoint reads distances several times too far at this camera height, which we found by comparing predicted with ground-truth depth. CARLA supplies oracle full-extent boxes, so these results isolate the depth mechanism from detector behavior; the KITTI analysis of Appendix~\ref{app:mechanism} covers the detector.

\textbf{Results.} The score reaches AUC 0.819 [0.801, 0.834] (frame bootstrap) against 0.644 for distance alone, and beats distance in every band (Table~\ref{tab:carla}). Performance falls with range because the indoor checkpoint saturates beyond about 8\,m, and at this camera height its depth error has a substantial offset, the case Eq.~\ref{eq:affine} warns about. The KITTI and CARLA results have complementary weaknesses: KITTI has real imagery but coarse human labels, CARLA has exact labels but synthetic imagery and oracle boxes.

\begin{table}[htbp]
\caption{Occlusion score on CARLA (visibility $<$ 0.8 vs.\ $\geq$ 0.8), by distance band. $N$ counts pedestrians; intervals bootstrap frames.}
\label{tab:carla}
\centering
\small
\begin{tabular}{lccc}
\toprule
Distance band & $N$ & AUC of score [95\% CI] & AUC of distance \\
\midrule
0--5\,m   & 154  & 0.969 [0.939, 0.989] & 0.506 \\
5--10\,m  & 668  & 0.895 [0.868, 0.919] & 0.585 \\
10--15\,m & 687  & 0.839 [0.804, 0.871] & 0.562 \\
15--25\,m & 1277 & 0.797 [0.771, 0.821] & 0.509 \\
25--60\,m & 507  & 0.739 [0.693, 0.783] & 0.481 \\
all       & 3293 & \textbf{0.819 [0.801, 0.834]} & 0.644 \\
\bottomrule
\end{tabular}
\end{table}

\section{Runtime}
\label{app:runtime}

Table~\ref{tab:runtime} reports median per-frame latency on 40 KITTI test frames on a laptop CPU (Intel Core i7-8665U, 4 threads, eager PyTorch, fp32, no GPU), with a YOLOv8n detector at 640\,px and Depth Anything V2 Metric-Outdoor-Small. The depth network keeps the frame's aspect ratio, so a 518\,px short side is a 518$\times$1722 input on KITTI. The score itself, a box percentile and a product, costs about 0.3\,ms; the depth network is 94--98\% of the pipeline. A smaller depth input is about 5$\times$ faster, but on this small sample the resulting scores rank-correlate with the full-size ones at only 0.68, so its occlusion AUC is untested. We have not measured embedded hardware, compiled or quantized depth networks, or lighter depth models, and make no real-time claim.

\begin{table}[htbp]
\caption{Median CPU latency per frame (ms), YOLOv8n + Depth Anything V2 Metric-Outdoor-Small.}
\label{tab:runtime}
\centering
\small
\begin{tabular}{lccccc}
\toprule
Depth input & Detection & Depth & Score & Total & FPS \\
\midrule
518$\times$1722 (default) & 77 & 6233 & 0.3 & 6336 & 0.16 \\
294$\times$980            & 66 & 1229 & 0.3 & 1314 & 0.76 \\
\bottomrule
\end{tabular}
\end{table}

\section{Implementation and Evaluation Details}
\label{app:implementation}

KITTI is split 5{,}236/1{,}122/1{,}123 and CityPersons 2{,}432/521/522 for train/validation/test. Three seeds are used for every configuration where a confidence interval is reported; single-seed results are marked $\dagger$ in Table~\ref{tab:elim}, giving 81 evaluated checkpoints in total. All models train with SGD (lr0 0.0013, lrf 0.123, weight decay 0.0054) for 150 epochs, batch size 32, patience 20, at 640\,px. Occlusion augmentation is applied only during training, and only when more than half of the ground-truth pedestrian remains visible. Depth Anything V2's metric output is clipped to 0--80\,m. The occlusion score is evaluated on 1{,}941 matched KITTI test detections (682 pedestrians, three YOLOv8x seeds); AUC treats ties as one half, and confidence intervals use 2{,}000 bootstrap resamples of pedestrians (of frames on CARLA). Scripts reproducing every number in the paper are released with it.

\section{Limitations}
\label{app:limitations}

\textbf{What is validated.} The signal is validated as a ranking of occlusion presence and a coarse category (partly, heavily) against KITTI's discrete labels, and against one visibility threshold on CARLA; it is not a continuous estimator of occlusion fraction. Evaluation covers pedestrians in daytime imagery only, not other object classes, whose size is far less standardized than human stature.

\textbf{Detector dependence.} The apparent-height assumption holds for predicted boxes at the median but fails in a tail of low-confidence, heavily occluded detections, and the detector misses 27\% of largely occluded pedestrians outright. A detector that systematically shrinks occluded boxes would weaken the signal, and should be checked the same way before deployment.

\textbf{Depth model and camera.} Calibration-free ranking holds only when depth error is close to a pure gain over the operating range. It held on KITTI with the outdoor checkpoint, but not with the indoor checkpoint at robot height on CARLA. On our own real-road robot recording, with the outdoor checkpoint at 0.186\,m, a thresholded version of the score did no better at its operating point than flagging the smallest boxes. A new depth model, mounting height, or set of camera intrinsics must be checked, for example with the label-free correlation of Appendix~\ref{app:formulation}, not assumed.

\textbf{Untested conditions.} We have not evaluated the score in dense crowds, where several people share one box's nearest surface; at night or in rain; with other monocular depth models; or on embedded hardware. The score cannot distinguish an occluder from any other nearer surface inside the box. The Lagos evaluation measures detector transfer, not the score, because the Lagos annotations have no occlusion labels; labelling occlusion there is the most direct next test of the claim this paper is motivated by.

\end{document}